\pdfoutput=1
\documentclass[11pt]{article}
\usepackage{emnlp2017}
\usepackage{times}
\usepackage{url}
\usepackage{latexsym}
\usepackage{graphicx}
\usepackage{enumitem}
\usepackage{amssymb}
\usepackage{amsmath}
\usepackage{color}
\usepackage{xfrac}
\usepackage{lipsum}
\usepackage{soul}
\usepackage{multirow}

\newcommand{\myfunctionname}[1]{{\operatorname{\mathit{#1}}}} 

\usepackage[british]{babel} 
\usepackage[utf8]{inputenc}
\usepackage{color}
\input epsf

\newcommand{\changed}[1]{\textcolor{black}{#1}}
\emnlpfinalcopy 

\title{Towards Syntactic Iberian Polarity Classification 
\thanks{\ \ DV was funded by MECD (FPU13/01180). MG is funded by a \emph{Juan de la Cierva} grant (FJCI-2014-22853). CGR has received funding from the ERC, under the European Union's Horizon 2020 research and innovation programme (FASTPARSE, grant agreement No 714150). This research was supported by MINECO (FFI2014-51978-C2).}
}

\author{David Vilares$^\spadesuit$$^\dag$, Marcos Garcia$^\spadesuit$$^\ddag$, Miguel A. Alonso$^\spadesuit$$^\dag$, Carlos Gómez-Rodríguez$^\clubsuit$$^\spadesuit$$^\dag$ \\
  Universidade da Coru\~{n}a \\
  $^\clubsuit$FASTPARSE Lab, $^\spadesuit$LyS Group  \\
  $^\dag$ Departamento de Computaci\'{o}n, Campus de Elvi\~{n}a \\
  $^\ddag$Departamento de Letras, Campus da Zapateira \\ 
  15701, A Coru\~{n}a, Spain \\
  {\tt david.vilares@udc.es,marcos.garcia.gonzalez@udc.es} \\
  {\tt miguel.alonso@udc.es,carlos.gomez@udc.es} 
}

\date{}

\begin{document}
\maketitle

\begin{abstract}
Lexicon-based methods using syntactic rules for polarity classification rely on parsers that are dependent on the language and on  treebank guidelines. Thus, rules are also dependent and require adaptation, especially in multilingual scenarios. We tackle this challenge in the context of the Iberian Peninsula, releasing the first symbolic syntax-based Iberian system with rules shared across five official languages: Basque, Catalan, Galician, Portuguese and Spanish. The model is made available.\footnote{The resources used in this work have been integrated as a part of \url{https://github.com/aghie/uuusa}}  
\end{abstract}

\section{Introduction}

Finding the scope of linguistic phenomena in natural language processing ({\sc nlp}) is a core utility of parsing. In sentiment analysis ({\sc sa}), it is used to address structures that play a role in polarity classification, both in supervised \cite{SocPerWuChuManNgPot2013a} and symbolic \cite{VilAloGom2015} models. In the latter case, these are mostly monolingual and dependent on the annotation of the training treebank, and so the rules are annotation-dependent too. Advances in {\sc nlp} make it now possible to overcome such issues. We present a model that analyzes five official languages in the Iberian Peninsula: Basque (\emph{eu}), Catalan (\emph{ca}), Galician (\emph{gl}), Portuguese (\emph{pt}) and Spanish (\emph{es}). We rely on three premises:

\begin{enumerate}[leftmargin=0cm,itemindent=.5cm,labelwidth=\itemindent,labelsep=0cm,align=left]
\itemsep-0.1em
\item Syntactic structures can be defined in a universal way \cite{nivre2015universal}\label{premise-1}.
\item Training a single model for multilingual parsing is feasible \cite{ammar2016many}. 
\item We can define universal rules for various phenomena, if \ref{premise-1} is assured \cite{VilGomAloArxiv2016}.
\end{enumerate}

Based on those, we: (a) combine existing subjectivity lexica, (b) train an \emph{Iberian} tagger and parser, and (c) define a set of Iberian syntax-based rules. The main contributions of the paper are: 

\begin{enumerate}[leftmargin=0cm,itemindent=.5cm,labelwidth=\itemindent,labelsep=0cm,align=left]
\itemsep-0.1em
\item A single set of syntactic rules to handle linguistic phenomena across five Iberian languages from different families. 
%
%
%
%
%

\item The first end-to-end multilingual syntax-based {\sc sa} system that analyzes five official languages of the Iberian Peninsula.
%
%
This is also the first evaluation for {\sc sa} that provides results for some of them.

\end{enumerate}

\section{Related work}

Polarity classification has been addressed through machine learning  \cite{MohKirZhu2013a,SocPerWuChuManNgPot2013a,vo-zhang:2016:P16-2}, and lexicon-based models \cite{Turney}. Most of the research involves English texts, although studies can be found for other languages such as Chinese \cite{chen2016implicit} or Arabic \cite{shoukry2012sentence}.

For the official languages in the Iberian Peninsula, much of the literature has focused on Spanish. \newcite{BroTofTab2009a} proposed a lexicon-based {\sc sa} system that defines rules at the lexical level to handle negation, intensification or adversative subordinate clauses. They followed a cross-lingual approach, adapting their English method \cite{Lexicon-BasedMethods} to obtain the semantic orientation ({\sc so}) of Spanish texts.  \newcite{VilAloGom2015} created a syntactic rule-based system, by making an interpretation of Brooke et al.'s system, but limited to AnCora trees \cite{Ancora}. \newcite{martinez2011opinion} were one of the first to report a wide set of experiments on a number of bag-of-words supervised classifiers. The TASS workshop on sentiment analysis focused on Spanish language \cite{TASS-2012} annually proposes different challenges related to polarity classification, and a number of approaches have used its framework to build their Spanish systems, most of them based 
on supervised learning \cite{urizar2013elhuyar,GamGarFer2013a,hurtado2015elirf,VilAloGom2015d}.  

Sentiment analysis for Portuguese has also attracted the interest of the research community. \newcite{silva2009design} presented a system for detection of opinions about Portuguese politicians. \newcite{souza2011construction} built a lexicon for Brazilian Portuguese exploring different techniques (e.g. translation and thesaurus-based approaches) and available resources. \newcite{souza2012sentiment} carried out a study of Twitter data, exploring preprocessing techniques, subjectivity data and negation approaches. They concluded that those have a small impact on the polarity classification of tweets.  \newcite{balage2013evaluation} evaluate the quality of the Brazilian {\sc liwc} dictionary \cite{PennebakerFrancisBooth2001a} for {\sc sa}, comparing it with existing lexica for this language.

For Basque, Catalan and Galician, literature is scarce. \newcite{Senticon} introduce a method to create multiple layered lexicons for different languages including co-official languages in Spain. \newcite{SANVICENTE16.468} explore different ways to create lexicons, and apply them to the Basque case. They report an evaluation on a Basque dataset intended for polarity classification. \newcite{BoscoCatalanDebates} discuss the collection of data for the Catalan Elections and design an annotation scheme to apply {\sc sa} techniques, but the dataset is still not available. 
With respect to Galician, in this article we will present the first published results for this language.

\section{\label{sec:sisa}SISA: Syntactic Iberian SA}

\subsection{Preliminaries}

\newcite{VilGomAloArxiv2016} propose a formalism to define \emph{compositional operations}. Given a dependency tree for a text, a compositional operation defines how a node in the tree modifies the semantic orientation ({\sc so}) of a branch or node, based on elements such as the word form, part-of-speech (PoS) tag or dependency type, without limitations in terms of its location inside such tree. They released an implementation, where an arbitrary number of practical compositional operations can be defined. The system queues and propagates them through the tree, until the moment they must be dequeued and applied to their target.
The authors showed how the same set of operations, defined to work under the Universal Treebank ({\sc ut}) guidelines \cite{Mcdonald2013}, can be shared across languages, but they do not explore how to create a single pipeline for analyzing many languages. This paper explores that path in the context of Iberian Peninsula, presenting an unified syntactic Iberian {\sc sa} model ({\sc sisa}).

We below present how to build {\sc sisa}, from the bottom (subjectivity lexica, tagging and dependency parsing) to the top levels (application of compositional operations to compute the final {\sc so}).

\subsection{Subjectivity Lexica}
{\sc sisa} needs multilingual polarity lexica in order to predict the sentiment of a text.  
We used two sets of monolingual lexica as our starting points:


\begin{enumerate}[leftmargin=0cm,itemindent=.5cm,labelwidth=\itemindent,labelsep=0cm,align=left]
\itemsep-0.1em
\item \emph{Spanish SFU lexicon} \cite{BroTofTab2009a}: It contains {\sc so}'s for subjective words that range from 1 to 5 for positive and negative terms. We translated it to \emph{ca}, \emph{eu}, \emph{gl } and \emph{pt} using \emph{apertium} \cite{forcada2011apertium}. 
We removed the unknown words and obtained the numbers in Table~\ref{tab:lex1}.\footnote{We used the original \emph{apertium} outputs, except for the \emph{pt} and \emph{gl} lexica (manually reviewed by a linguist).} 

\item \emph{ML-Senticon} \cite{Senticon}: Multi-layered lexica (not available for \emph{pt}) with {\sc so}'s where each layer contains a larger number of terms, but less trustable. We used the seventh layer for each language. As \emph{eu}, \emph{ca} and \emph{gl} files have the same PoS-tag for adverbs and adjectives, they were automatically classified using monolingual tools \cite{ixa,freeling,citius} (Table~\ref{tab:lex2} contains the statistics).
{\sc so}'s (originally from 0 to 1) were linearly transformed to the scale of the {\sc sfu} lexicon. 
\end{enumerate}




\begin{table}
\centering
\tabcolsep=0.17cm
\small{
  \begin{tabular}{|c|rrrrr|}
  \hline
    \bf Tag  &\bf es &\bf pt &\bf ca &\bf eu &\bf gl\\ \hline \hline
    {\sc adj}  &   2,045 &      1,865 &   1,686 &  1,757 &    2,002\\
    {\sc noun} &   1,323 &      1,183 &   1,168 &  1,211 &    1,270\\
    {\sc adv}  &     594 &        570 &     533 &    535 &      599\\
    {\sc verb} &     739 &        688 &     689 &    563 &      723\\
    \hline
  \end{tabular}}
  \caption{\label{tab:lex1}Size of the SFU (single words) lexica.}
\end{table}

\begin{table}
\centering
\small{
  \begin{tabular}{|c|rrrr|}
  \hline
    \bf Tag  &\bf es & \bf ca & \bf eu & \bf gl \\ \hline \hline
    {\sc adj}  &   2,558 &   1,619 &     22 &    1,530 \\
    {\sc noun} &   2,094 &   1,535 &  1,365 &      579 \\
    {\sc adv}  &     117 &     23  &      3 &       26 \\
    {\sc verb} &     603 &     500 &    272 &      144 \\
    \hline
  \end{tabular}}
  \caption{\label{tab:lex2}Size of the resulting ML-Senticon lexica.}
\end{table}

The SFU and ML-Senticon lexica for each language were combined to obtain larger monolingual resources, and these were in turn combined into a common Iberian lexicon 
(see Table~\ref{tab:lex3}). When merging lexica, we must consider that: 

\begin{table}
\centering
\tabcolsep=0.13cm
\small{
  \begin{tabular}{|l|rrrrrr|}
    \hline
    \bf Tag  &\bf es &\bf pt & \bf ca & \bf eu & \bf gl & \bf Iberian \\ \hline \hline
    {\sc adj} &   3,775 & 1,865 &  2,704 &  1,529 &    2,990 &       9,385  \\
    {\sc noun} &   3,079 & 1,183 &  2,377 &  2,392 &    1,684 &       8,733  \\
    {\sc adv}  &     665 & 570 &   545 &    485 &      612 &       1,891  \\
    {\sc verb} &   1,177 & 688 & 1,034 &    728 &      801 &       2,998  \\
    \hline
  \end{tabular}}
  \caption{\label{tab:lex3}Size of the final lexica.} 
\end{table}

\begin{enumerate}[leftmargin=0cm,itemindent=.5cm,labelwidth=\itemindent,labelsep=0cm,align=left]
\itemsep-0.1em
\item In monolingual mergings, the same word can have different {\sc so}'s. E.g., the Catalan adjective  \textquoteleft abandonat\textquoteright\ (\emph{abandoned}) has $-1.875$ and $-3$ in ML-Senticon and SFU, respectively.
%
%
%

\item When combining lexica of different languages, the same word form might have different meanings (and \textsc{so}s) in each language. Merging them in a multilingual resource could be problematic. For example, the adjective \textquoteleft espantoso\textquoteright\ has a value of $-4.1075$ in the combined \emph{es} lexicon (\emph{frightening}), and of $-3.125$ in the \emph{gl} one (\emph{frightening}), while the same word in the \emph{pt} data (\emph{astonishing}) has a positive value of $5$. Note, however, that even if they could be considered very similar from a lexical or morphological perspective, many \changed{phonological false friends} have different spellings in each language (such as the negative \textquoteleft vessar\textquoteright\ (\emph{to spill}) in \emph{ca} and the positive \textquoteleft besar\textquoteright\ (\emph{to kiss}) in \emph{es}), so these cases end up not being a frequent problem (only $0.36\%$ of the words have both positive and negative polarity in the monolingual lexica).

\end{enumerate}

These two problems were tackled by averaging the polarities of words with the same form. Thus, the first monolingual mergings produced a balanced {\sc so} (e.g., `abandonat' has $-2.4375$ in the combined \emph{ca} lexicon), while in the subsequent multilingual fusion, contradictory false friends have a final value close to \emph{no polarity} (e.g., `espantoso', with a {\sc so} of $-0.7$ in the Iberian lexicon). The impact of these mergings is analyzed in \S \ref{section-evaluation}.

\subsection{PoS-tagging and dependency parsing}

For the compositional operations to be triggered, we first need to do the tagging and the dependency parse for a sentence. To do so, we trained an Iberian PoS-tagger and parser, i.e. single modules that can analyze Iberian languages without applying any language identification tool. Multilingual taggers and parsers can be trained following approaches based on  \cite{VilGomAloACL2016,ammar2016many}.  We are relying on the Universal Dependency ({\sc ud}) guidelines \cite{nivre2015universal} to train these tools, since they provide corpora for all languages studied in this paper.

For the Iberian tagger we relied on \newcite{TouMan2000a}, obtaining the following accuracies (\%) in the monolingual {\sc ud} test sets: \emph{pt} (95.96), \emph{es} (94.37), \emph{ca} (97.41), \emph{eu} (93.88) and \emph{gl} (94.09). For the Iberian parser we used the approach by \newcite{VilGomAloACL2016}, whose performance ({\sc las/uas)\footnote{{\sc las/uas}: The percentage of arcs where both the head and dependency type / the head are correct.}} on the same {\sc ud} test sets was: \emph{pt} (78.78/84.50), \emph{es} (80.20/85.23), \emph{cat} (84.01/88.08), \emph{eu} (62.01/71.64)\footnote{The parsing results for Basque (with a high proportion of non-projective trees) were worse than expected. However, the parser trained based on the method by \newcite{VilGomAloACL2016}  automatically selected a projective algorithm for training, as the average prevalence of non-projectivity across our five Iberian languages is low. We hypothesize that this is the main reason of the lower performance for this language.} and \emph{gl} (75.65/82.11).



\subsection{Compositional operations}

For a detailed explanation of compositional operations, we encourage the reader to consult \newcite{VilGomAloArxiv2016}, but we here include an overview as part of \textsc{sisa}. Briefly, a compositional operation is tuple $o=(\tau,C,\delta,\pi,S)$ such that:
\begin{itemize}
\itemsep-0.1em
\item $\tau : \mathbb{R} \rightarrow \mathbb{R}$ is a  transformation function to apply on the semantic orientation of nodes, where $\tau$ can be $\myfunctionname{weighting}_{\beta}(SO) = SO \times (1+\beta)$ or $\myfunctionname{shift}_{\alpha}(SO) = \left\{
	\begin{array}{ll}
		SO - \alpha  & \mbox{if } SO \geqq 0 \\
		SO + \alpha & \mbox{if } SO <0 \\
	\end{array}
\right.$,
\item $C : V \rightarrow \{true,false\}$ is a predicate that determines whether a node in the tree will trigger the operation, based on word forms, PoS-tags and dependency types,
\item \changed{$\delta \in \mathbb{N}$ is a number of levels that we need to ascend in the tree to calculate the scope of $o$, i.e., the nodes of\, $T$ whose SO is affected by the transformation function $\tau$,}
\item $\pi$ is a priority used to break ties when several operations coincide on a given node, and
\item $S$ is a scope function that will be used to determine the nodes affected by the operation.
\end{itemize}

\changed{We adapt the {\sc ut} operations used by \newcite{VilGomAloArxiv2016} to the {\sc ud} style to handle, which are now described:}
\begin{enumerate}[leftmargin=0cm,itemindent=.5cm,labelwidth=\itemindent,labelsep=0cm,align=left]
\itemsep-0.1em
	\item \emph{Intensification}: It diminishes or amplifies the {\sc so} of a word or a phrase. It operates from adjectives or adverbs modifying the {\sc so} of the head structure they depend on: e.g., the {\sc so} of `grande' (\emph{big}, in \emph{es}) increases from $1.87$ to $2.34$ if a word such as `muy' (\emph{very}) depends on it and its labeled with the dependency type \emph{advmod}. Formally, for o$_{intensification}$, $\tau = weight_\beta(SO)$, $C = {w \in {\text{\small{intensifiers}}}} \wedge {t \in \{\text{\small{ADV,ADJ}}\}} \wedge {d \in \{\text{\small{advmod,amod,nmod}}\}} $, $\delta=1$, $\pi = 3$ and $S = $ \{\emph{target node, b(advmod), b(amod)}\}, where \emph{b(x)} indicates that the scope is the first branch at the target level whose dependency type is $x$.  $\beta$ is extracted from a lexicon with \emph{booster} values (in this work obtained from {\sc sfu}, where `muy' has a \emph{booster} value of $0.25$).
    


    \item \emph{Subordinate adversative clauses}: This rule is designed for dealing with structures coordinated by adversative conjunctions (such as \emph{but}), which usually involve opposite polarities between the two joint elements (e.g., ``good but expensive"). Here, the {\sc so} of the first element is multiplied by $1-0.25$, so its polarity decreases. Formally, $\tau = weight_{-0.25}(SO)$, $C = {w \in {\text{\small{adversatives}}}} \wedge t \in \{\text{\small{CONJ,SCONJ}}\} \wedge {d \in \{\text{\small{cc,advmod,mark}}\}} $, $\delta=1$, $\pi = 1$ and $S = $ \{\emph{subjl}\}. \emph{Subjl} indicates that the scope is the first left branch with SO $!=0$ at the target level.

    \item \emph{Negation}: In most cases, negative adverbs shift the polarity of the structures they depend on (``It is nice" \emph{versus} ``It is not nice"). 
    In order to handle these cases, the present rule shifts the polarity of the head structures of a negative adverb by $\alpha$ (where $\alpha=4$, in our experiments). In the previous example, the polarity of ``nice'' would drop from $3.5$ to $-0.5$ if affected by the rule. Formally, for o$_{negation}$, $\tau = shift_{4}(SO)$, $C = {w \in {\text{\small{negators}}}} \wedge {d \in \{\text{\small{neg,advmod}}\}} $, $\delta=1$, $\pi = 2$ and $S = $ \{\emph{target node, b(root), b(cop), b(nsubj), subjr, all}\}. \emph{Subjr} indicates that the scope is the first branch with SO $!=0$ and \emph{all} indicates to apply negation at the target level as a backoff option, if none of the previous scopes matched.

    \item \emph{\textquoteleft If\textquoteright\ irrealis}: In conditional statements, a {\sc sa} system may obtain an incorrect polarity due to the presence of polarity words which actually do not reflect a real situation (``This is good" \emph{vs} ``If this is good"). This rule attempts to better analyze these structures by shifting the polarity (here, multiplied by $-1$) if a conditional conjunction depends on it. Formally, for o$_{irrealis}$, $\tau = weight_{-1}(SO)$, $C = {w \in {\text{\small{irrealis}}}} \wedge {d \in \{\text{\small{mark,advmod,cc}}\}} $, $\delta=1$, $\pi = 3$ and $S = $ \{\emph{target node, subjr}\}.
    

\end{enumerate}


\section{Evaluation}\label{section-evaluation}

This section presents the results of the experiments we carried out
with our system using both the monolingual and the multilingual lexica,
compared to the performance of a supervised classifier for three of the
five analyzed languages. 

\subsection{Testing corpora}

\begin{itemize}[leftmargin=0cm,itemindent=.5cm,labelwidth=\itemindent,labelsep=0cm,align=left]
\itemsep-0.1em
\item \emph{Spanish SFU} \cite{BroTofTab2009a}: A set of 400 long reviews (200 positive, 200 negative) from different domains such as movies, music, computers or washing machines.

\item \emph{Portuguese SentiCorpus-PT 0.1} \cite{senticorpus-pt}: A collection of comments from the Portuguese newspaper \emph{Público} with polarity annotation at the entity level. As our system assigns the polarity at the sentence level, we selected the SentiCorpus sentences with (a) only one \textsc{so} and (b) with $>1$ \textsc{so} iff all of them were the same, generating a corpus with $2,086$ (from $2,604$) sentences.

\item \emph{Basque Opinion Dataset} \cite{SANVICENTE16.468}: Two small corpora in Basque containing news articles and reviews (music and movie domains).  We merged them to create a larger dataset, containing a total of 224 reviews.
%
%
%
%

\item[]\noindent In addition, due to the lack of available sentence- or document-level corpora for Catalan or Galician, we opted for synthetic corpora:

\item \emph{Synthetic Catalan SFU}: An automatically translated version to \emph{ca} of the Spanish SFU, with 5\% of the words from the original corpus considered as unknown by the translation tool.
\item \emph{Synthetic Galician SFU}: An automatically translated version to \emph{gl} of the Spanish SFU ($\approx6.4\%$ of the words not translated).
%
%
%

\end{itemize}

\subsection{Experiments}

We performed different experiments on binary polarity classification for knowing (a) the accuracy of the system, (b) the impact of the merged resources, and (c) the impact of the universal rules in monolingual and multilingual settings:

\begin{enumerate}[leftmargin=0cm,itemindent=.5cm,labelwidth=\itemindent,labelsep=0cm,align=left]
\itemsep-0.1em
\item \emph{SL-O}: Single lexica, no operations (baseline).
\item \emph{ML-O}: Multilingual lexica, no operations. \label{configuration-multilingual-lexica-no-rules}
\item \emph{SL+O}: Single lexica with universal operations.
\item \emph{ML+O}: Multilingual lexica with universal operations.
\end{enumerate}

The performance of our system was compared to LinguaKit (\emph{LKit}), an open-source toolkit which performs supervised sentiment analysis in several languages \cite{GamGarFer2013a,linguakit}.


\begin{table}
\centering
\tabcolsep=0.07cm
\small{
  \begin{tabular}{|l|r@{\ \ }r@{\ \ }r@{\ \ }r@{\ \ }r|}
    \hline
    \bf Lg & \bf SL-O & \bf SL+O & \bf ML-O & \bf ML+O & \bf LKit \\
    \hline \hline
    es       &     60.00 &   75.75 &     63.75 & \textbf{76.50} &  58.75 \\
    ca      &     54.00 &   57.50 &     58.25 & \textbf{73.00} &   --- \\
    gl       &     60.75 &   \textbf{73.00} &     60.00 & 70.00 & 50.25 \\
    eu      &     62.95 &   69.20 &     65.63 & \textbf{72.32} &  --- \\
    pt       &     60.50 &   \textbf{67.35} &     57.29 & 65.01 & 60.55 \\
    \hline
  \end{tabular}}
  \caption{\label{tab:results}Results of the different tests. In \emph{LKit} we only evaluated the positive and negative results (it also classifies sentences with no polarity).}
\end{table}

Table~\ref{tab:results} shows the results of each of these models on the different corpora. The baseline (\emph{SL-O}) obtained values between $54\%$ (\emph{ca}) and $62.95\%$ (\emph{eu}), \changed{results that are in line to those obtained by the supervised model.}\footnote{LinguaKit was intended for tweets (not long texts).}
As we are not aware of available {\sc sa} tools for \emph{ca}, we could not compare our results with other systems. For Basque, \newcite{SANVICENTE16.468} evaluated several lexica (both automatically translated and extracted, as well as with human annotation) in the same dataset used in this paper. They used a simple average polarity ratio classifier, which is similar to our baseline. Even if the lexica are different, their results are very similar to our \emph{SL-O} system ($63\%$ \emph{vs} $62,95\%$), and they also show that manually reviewing the lexica can boost the accuracy by up to $13$\%. 

%
%

The central columns of Table~\ref{tab:results} show the results of using universal rules and a merged lexicon in the same datasets. In \emph{gl} and \emph{pt} the best values were obtained using individual lexica together with syntactic rules, while the Iberian system achieved the best results in the other languages.

\begin{table}
\centering
\tabcolsep=0.07cm
\small{
  \begin{tabular}{|l|r@{\ \ }r@{\ \ }|r@{\ \ }r|}
    \hline
    \bf Lg & \bf O(SL) & \bf O(ML) & \bf ML(-O) & \bf ML(+O)\\
    \hline \hline
    es       &     15.75 &     12.75 &         3.75 &       0.75  \\
    ca      &      3.50 &     14.75 &         4.25 &       15.5  \\
    gl       &     12.25 &     10.00 &        -0.75 &      -3.00  \\
    eu      &      6.25 &      6.69 &         2.68 &       3.12  \\
    pt       &      6.85 &      7.72 &        -3.21 &      -2.34  \\
    \hline
  \end{tabular}}
  \caption{\label{tab:impact}Impact of the operations (O) with mono ({\sc sl}) and multilingual lexica ({\sc ml}) and of the {\sc ml} with (+O) and without operations (-O).}
\end{table}

Table~\ref{tab:impact} summarizes the impact that the rules have in both the monolingual and the multilingual setting, as well as the differences in performance due to the fusion process. 
Concerning the rules (columns 2 and 3), the results show that using the same set of universal rules improves the performance of the classifier in all the languages and settings. 
Their impact varies between 
$3.5$ percentage points (\emph{ca}) and more than $15$ (\emph{es}) and, for each language, the rules provide a similar effect in monolingual and multilingual lexica (except for \emph{ca}, with much higher values in the ML scenario).

The fusion of the different lexica had different results (columns 4 and 5 of Table~\ref{tab:impact}): in \emph{gl} and \emph{pt}, it had a negative impact (between $-0.75\%$ and $-3.21\%$) while in the other three the ML setting achieved better values (between $0.75$ and $15.5$ points, again with huge differences in \emph{ca}). On average, using multilingual lexica had a positive impact of $1.3$ (-O) and $2.8$ points (+O).
As mentioned, \emph{ca} has a different behaviour: the gain from rules when using monolingual lexica is about $3.50$ points (lower than other languages), and the benefit of the ML lexicon without syntactic rules is of $4.25$ points. However, when combining both the universal rules and the ML lexicon its performance increases $\approx 15$ points, turning out that the combination of these two factors is decisive.


In sum, the results of the experiments indicate that syntactic rules defined by means of a harmonized annotation can be used in several languages with positive results. Furthermore, the merging of monolingual lexica (some of them automatically translated) can be applied to perform multilingual {\sc sa} with little impact in performance when compared to language-dependent systems.

\section{Conclusions and current work}

We built a single symbolic syntactic system for polarity classification that analyzes five official languages of the Iberian peninsula. With little effort we obtain robust results for many languages. 
As current work, we are working on texts harder to parse and low-resource languages: we developed a Galician corpus of manually labeled tweets, 
%
%
where {\sc sisa} obtains between 62\% and 65\% accuracy for different settings,\footnote{This corpus is available at \url{http://grupolys.org/software/CHIOS-SISA/}} and plan to incorporate \newcite{KonSchSwaBhaDyeSmi2014a} parser to improve its performance.

\bibliographystyle{emnlp_natbib}
\bibliography{wassa2017.bib}

\begin{thebibliography}{37}
\expandafter\ifx\csname natexlab\endcsname\relax\def\natexlab#1{#1}\fi

\bibitem[{Agerri et~al.(2014)Agerri, Bermudez, and Rigau}]{ixa}
R.~Agerri, J.~Bermudez, and G.~Rigau. 2014.
\newblock \href
  {http://www.lrec-conf.org/proceedings/lrec2014/pdf/775_Paper.pdf} {{IXA
  pipeline: Efficient and Ready to Use Multilingual NLP tools}}.
\newblock In \emph{Proceedings of the 9th edition of the Language Resources and
  Evaluation Conference (LREC 2014)}, pages 3823--3828.

\bibitem[{Ammar et~al.(2016)Ammar, Mulcaire, Ballesteros, Dyer, and
  Smith}]{ammar2016many}
Waleed Ammar, George Mulcaire, Miguel Ballesteros, Chris Dyer, and Noah Smith.
  2016.
\newblock \href {https://transacl.org/ojs/index.php/tacl/article/view/892}
  {Many languages, one parser}.
\newblock \emph{Transactions of the Association for Computational Linguistics},
  4:431--444.

\bibitem[{Balage~Filho et~al.(2013)Balage~Filho, Pardo, and
  Alu\'{i}sio}]{balage2013evaluation}
P.~P. Balage~Filho, T.~AS Pardo, and S.~M. Alu\'{i}sio. 2013.
\newblock \href {http://www.aclweb.org/anthology/W13-4829} {{An evaluation of
  the Brazilian Portuguese LIWC dictionary for sentiment analysis}}.
\newblock In \emph{Proceedings of the 9th Brazilian Symposium in Information
  and Human Language Technology (STIL)}, pages 215--219.

\bibitem[{Bosco et~al.(2016)Bosco, Lai, Patti, Rangel~Pardo, and
  Rosso}]{BoscoCatalanDebates}
C.~Bosco, M.~Lai, V.~Patti, F.~M. Rangel~Pardo, and P~Rosso. 2016.
\newblock Tweeting in the debate about catalan elections.
\newblock In \emph{Proceedings of the Tenth International Conference on
  Language Resources and Evaluation (LREC 2016). Emotion and Sentiment Analysis
  Workshop.}, pages 67--70.

\bibitem[{Brooke et~al.(2009)Brooke, Tofiloski, and Taboada}]{BroTofTab2009a}
J.~Brooke, M.~Tofiloski, and M.~Taboada. 2009.
\newblock \href {http://anthology.aclweb.org/R/R09/R09-1.pdf#page=74}
  {{Cross-Linguistic Sentiment Analysis: From English to Spanish}}.
\newblock In \emph{Proceedings of RANLP 2009, Recent Advances in Natural
  Language Processing}, pages 50--54, Bovorets, Bulgaria.

\bibitem[{Carvalho et~al.(2011)Carvalho, Sarmento, Teixeira, and
  Silva}]{senticorpus-pt}
P.~Carvalho, L.~Sarmento, J.~Teixeira, and M.~J. Silva. 2011.
\newblock \href {http://www.anthology.aclweb.org/P/P11/P11-2.pdf#page=604}
  {Liars and saviors in a sentiment annotated corpus of comments to political
  debates}.
\newblock In \emph{Proceedings of the 49th Annual Meeting of the Association
  for Computational Linguistics: Human Language Technologies: short
  papers-Volume 2}, pages 564--568. Association for Computational Linguistics.

\bibitem[{Chen and Chen(2016)}]{chen2016implicit}
H.~Chen and H.~Chen. 2016.
\newblock \href {https://www.aclweb.org/anthology/P/P16/P16-2004.pdf}
  {{Implicit Polarity and Implicit Aspect Recognition in Opinion Mining}}.
\newblock In \emph{The 54th Annual Meeting of the Association for Computational
  Linguistics}, pages 20--25.

\bibitem[{Cruz et~al.(2014)Cruz, Troyano, Pontes, and Ortega}]{Senticon}
F.~L Cruz, J.~A Troyano, B.~Pontes, and F.~J. Ortega. 2014.
\newblock \href
  {http://journal.sepln.org/sepln/ojs/ojs/index.php/pln/article/view/5041}
  {{ML-SentiCon: Un lexic{\'o}n multiling{\"u}e de polaridades sem{\'a}nticas a
  nivel de lemas}}.
\newblock \emph{Procesamiento del Lenguaje Natural}, 53:113--120.

\bibitem[{Forcada et~al.(2011)Forcada, Ginest{\'\i}-Rosell, Nordfalk,
  O’Regan, Ortiz-Rojas, P{\'e}rez-Ortiz, S{\'a}nchez-Mart{\'\i}nez,
  Ram{\'\i}rez-S{\'a}nchez, and Tyers}]{forcada2011apertium}
M.~L Forcada, M.~Ginest{\'\i}-Rosell, J.~Nordfalk, J.~O’Regan,
  S.~Ortiz-Rojas, J.~A. P{\'e}rez-Ortiz, F.~S{\'a}nchez-Mart{\'\i}nez,
  G.~Ram{\'\i}rez-S{\'a}nchez, and F.~M. Tyers. 2011.
\newblock \href
  {https://link.springer.com/content/pdf/10.1007/s10590-011-9090-0.pdf}
  {Apertium: a free/open-source platform for rule-based machine translation}.
\newblock \emph{Machine translation}, 25(2):127--144.

\bibitem[{Gamallo and Garcia(2017)}]{linguakit}
P.~Gamallo and M.~Garcia. 2017.
\newblock \href
  {http://www.linguamatica.com/index.php/linguamatica/article/viewFile/v9n1p2/392}
  {{LinguaKit: uma ferramenta multilingue para a análise linguística e a
  extração de informação}}.
\newblock \emph{Linguamática}, 9(1):19--28.

\bibitem[{Gamallo et~al.(2013)Gamallo, Garc\'{i}a, and
  Fern\'{a}ndez~Lanza}]{GamGarFer2013a}
P.~Gamallo, M.~Garc\'{i}a, and S.~Fern\'{a}ndez~Lanza. 2013.
\newblock \href {http://www.anthology.aclweb.org/S/S14/S14-2.pdf#page=191}
  {{TASS: A Naive-Bayes strategy for sentiment analysis on Spanish tweets}}.
\newblock In \emph{XXIX Congreso de la Sociedad Espa\~{n}ola de Procesamiento
  de Lenguaje Natural (SEPLN 2013). TASS 2013 - Workshop on Sentiment Analysis
  at SEPLN 2013}, pages 126--132, Madrid, Spain.

\bibitem[{Garcia and Gamallo(2015)}]{citius}
M.~Garcia and P.~Gamallo. 2015.
\newblock \href {https://link.springer.com/chapter/10.1007/978-3-319-27653-3_7}
  {{Yet Another Suite of Multilingual NLP Tools}}.
\newblock In \emph{{Languages, Applications and Technologies. Communications in
  Computer and Information Science}}, volume 563, pages 65--75. Springer.

\bibitem[{Hurtado et~al.(2015)Hurtado, Pla, and Buscaldi}]{hurtado2015elirf}
L.~F. Hurtado, F.~Pla, and D.~Buscaldi. 2015.
\newblock \href {http://ceur-ws.org/Vol-1397/elirf_upv.pdf} {{ELiRF-UPV en TASS
  2015: Análisis de Sentimientos en Twitter}}.
\newblock In \emph{Proceedings of TASS 2015: Workshop on Sentiment Analysis at
  SEPLN}, pages 35--40.

\bibitem[{Kong et~al.(2014)Kong, Schneider, Swayamdipta, Bhatia, Dyer, and
  Smith}]{KonSchSwaBhaDyeSmi2014a}
L.~Kong, N.~Schneider, S.~Swayamdipta, A.~Bhatia, C.~Dyer, and N.~A. Smith.
  2014.
\newblock \href {http://www.aclweb.org/anthology/D14-1108} {{A Dependency
  Parser for Tweets}}.
\newblock In \emph{Proceedings of the 2014 Conference on Empirical Methods in
  Natural Language Processing (EMNLP)}, pages 1001--1012, Doha, Qatar. ACL.

\bibitem[{Mart{\'\i}nez-C{\'a}mara et~al.(2011)Mart{\'\i}nez-C{\'a}mara,
  Mart{\'\i}n-Valdivia, and Ure{\~n}a-L{\'o}pez}]{martinez2011opinion}
Eugenio Mart{\'\i}nez-C{\'a}mara, M~Teresa Mart{\'\i}n-Valdivia, and L~Alfonso
  Ure{\~n}a-L{\'o}pez. 2011.
\newblock \href
  {https://link.springer.com/chapter/10.1007/978-3-642-22327-3_17} {Opinion
  classification techniques applied to a spanish corpus}.
\newblock In \emph{International Conference on Application of Natural Language
  to Information Systems}, pages 169--176. Springer.

\bibitem[{McDonald et~al.(2013)McDonald, Nivre, Quirmbach-Brundage, Goldberg,
  Das, Ganchev, Hall, Petrov, Zhang, T{\"a}ckstr{\"o}m et~al.}]{Mcdonald2013}
R.~T McDonald, J.~Nivre, Y.~Quirmbach-Brundage, Y.~Goldberg, D.~Das,
  K.~Ganchev, K.~B. Hall, S.~Petrov, H.~Zhang, O.~T{\"a}ckstr{\"o}m, et~al.
  2013.
\newblock \href {https://www.aclweb.org/anthology/P13-2017} {{Universal
  Dependency Annotation for Multilingual Parsing}}.
\newblock In \emph{Proceedings of the 51st Annual Meeting of the Association
  for Computational Linguistics}, pages 92--97. Association for Computational
  Linguistics.

\bibitem[{Mohammad et~al.(2013)Mohammad, Kiritchenko, and Zhu}]{MohKirZhu2013a}
S.~M Mohammad, S.~Kiritchenko, and X.~Zhu. 2013.
\newblock \href
  {http://www.aclweb.org/website/old_anthology/S/S13/S13-2.pdf#page=357}
  {{NRC-Canada: Building the State-of-the-Art in Sentiment Analysis of
  Tweets}}.
\newblock In \emph{Proceedings of the seventh international workshop on
  Semantic Evaluation Exercises (SemEval-2013)}, Atlanta, Georgia, USA.

\bibitem[{Nivre et~al.(2015)Nivre, Agi{\'c}, Aranzabe, Asahara, Atutxa,
  Ballesteros, Bauer, Bengoetxea, Bhat, Bosco et~al.}]{nivre2015universal}
J.~Nivre, {\v{Z}}.~Agi{\'c}, M.~J. Aranzabe, M.~Asahara, A.~Atutxa,
  M.~Ballesteros, J.~Bauer, K.~Bengoetxea, R.~A. Bhat, C.~Bosco, et~al. 2015.
\newblock Universal dependencies 1.2.

\bibitem[{Padr{\'o} and Stanilovsky(2012)}]{freeling}
L.~Padr{\'o} and E.~Stanilovsky. 2012.
\newblock \href
  {http://www.lrec-conf.org/proceedings/lrec2012/pdf/430_Paper.pdf} {{Freeling
  3.0: Towards wider multilinguality}}.
\newblock In \emph{Proceedings of the 8th edition of the Language Resources and
  Evaluation Conference (LREC 2012)}, Istambul.

\bibitem[{Pennebaker et~al.(2001)Pennebaker, Francis, and
  Booth}]{PennebakerFrancisBooth2001a}
J.~W. Pennebaker, M.~E. Francis, and R.~J. Booth. 2001.
\newblock \href {http://www.depts.ttu.edu/psy/lusi/files/LIWCmanual.pdf}
  {{Linguistic inquiry and word count: {LIWC} 2001}}.
\newblock \emph{Mahway: Lawrence Erlbaum Associates}, page~71.

\bibitem[{San~Vicente and Saralegi(2016)}]{SANVICENTE16.468}
I.~San~Vicente and X.~Saralegi. 2016.
\newblock \href
  {http://www.lrec-conf.org/proceedings/lrec2016/pdf/468_Paper.pdf} {Polarity
  lexicon building: to what extent is the manual effort worth?}
\newblock In \emph{Proceedings of the Tenth International Conference on
  Language Resources and Evaluation (LREC 2016)}, Paris, France. European
  Language Resources Association (ELRA).

\bibitem[{Saralegi and San~Vicente(2013)}]{urizar2013elhuyar}
X.~Saralegi and I.~San~Vicente. 2013.
\newblock \href
  {http://www.sepln.org/workshops/tass/2013/papers/tass2013-submission3-Elhuyar.pdf}
  {Elhuyar at tass 2013}.
\newblock In \emph{Proceedings of the Workshop on Sentiment Analysis at SEPLN
  (TASS 2013)}, pages 143--150.

\bibitem[{Shoukry and Rafea(2012)}]{shoukry2012sentence}
A.~Shoukry and A.~Rafea. 2012.
\newblock \href {http://ieeexplore.ieee.org/document/6261103/} {{Sentence-level
  Arabic sentiment analysis}}.
\newblock In \emph{Collaboration Technologies and Systems (CTS), 2012
  International Conference on}, pages 546--550. IEEE.

\bibitem[{Silva et~al.(2009)Silva, Carvalho, Sarmento, de~Oliveira, and
  Magalhaes}]{silva2009design}
M.~J Silva, P.~Carvalho, L.~Sarmento, E.~de~Oliveira, and P.~Magalhaes. 2009.
\newblock \href
  {https://web.fe.up.pt/~niadr/PUBLICATIONS/2009/epia2009-OPTIMISM-submitted.pdf}
  {{The design of OPTIMISM, an opinion mining system for Portuguese politics}}.
\newblock \emph{New trends in artificial intelligence: Proceedings of EPIA},
  pages 12--15.

\bibitem[{Socher et~al.(2013)Socher, Perelygin, Wu, Chuang, Manning, Ng, and
  Potts}]{SocPerWuChuManNgPot2013a}
R.~Socher, A.~Perelygin, J.~Wu, J.~Chuang, C.~D Manning, A.~Ng, and C.~Potts.
  2013.
\newblock \href {http://www.aclweb.org/anthology/D13-1170} {{Recursive Deep
  Models for Semantic Compositionality Over a Sentiment Treebank}}.
\newblock In \emph{EMNLP 2013. 2013 Conference on Empirical Methods in Natural
  Language Processing. Proceedings of the Conference}, pages 1631--1642,
  Seattle, Washington, USA. ACL.

\bibitem[{Souza and Vieira(2012)}]{souza2012sentiment}
M.~Souza and R.~Vieira. 2012.
\newblock \href {http://dl.acm.org/citation.cfm?id=2261085} {{Sentiment
  analysis on twitter data for portuguese language}}.
\newblock In \emph{International Conference on Computational Processing of the
  Portuguese Language}, pages 241--247. Springer.

\bibitem[{Souza et~al.(2011)Souza, Vieira, Busetti, Chishman, Alves, and
  Others}]{souza2011construction}
M.~Souza, R.~Vieira, D.~Busetti, R.~Chishman, I.~M. Alves, and Others. 2011.
\newblock \href {http://www.aclweb.org/anthology/W11-4507} {{Construction of a
  portuguese opinion lexicon from multiple resources}}.
\newblock In \emph{8th Brazilian Symposium in Information and Human Language
  Technology}, pages 59--66.

\bibitem[{Taboada et~al.(2011)Taboada, Brooke, Tofiloski, Voll, and
  Stede}]{Lexicon-BasedMethods}
M.~Taboada, J.~Brooke, M.~Tofiloski, K.~Voll, and M.~Stede. 2011.
\newblock \href {https://www.aclweb.org/anthology/J/J11/J11-2001.pdf}
  {{Lexicon-based methods for sentiment analysis}}.
\newblock \emph{Computational Linguistics}, 37(2):267--307.

\bibitem[{Taul{\'{e}} et~al.(2008)Taul{\'{e}}, Mart{\'{i}}, and
  Recasens}]{Ancora}
M.~Taul{\'{e}}, M.~A. Mart{\'{i}}, and M.~Recasens. 2008.
\newblock \href
  {http://www.lrec-conf.org/proceedings/lrec2008/pdf/35_paper.pdf} {{AnCora:
  Multilevel Annotated Corpora for Catalan and Spanish}}.
\newblock In \emph{Proceedings of the Sixth International Conference on
  Language Resources and Evaluation (LREC'08)}, pages 96--101, Marrakech,
  Morocco.

\bibitem[{Toutanova and Manning(2000)}]{TouMan2000a}
K.~Toutanova and C.~D. Manning. 2000.
\newblock \href {http://dl.acm.org/citation.cfm?id=1117802} {{Enriching the
  knowledge sources used in a maximum entropy part-of-speech tagger}}.
\newblock In \emph{Proceedings of the 2000 Joint SIGDAT conference on Empirical
  methods in natural language processing and very large corpora: held in
  conjunction with the 38th Annual Meeting of the Association for Computational
  Linguistics-Volume 13}, pages 63--70.

\bibitem[{Turney(2002)}]{Turney}
P.~D. Turney. 2002.
\newblock \href {https://doi.org/10.3115/1073083.1073153} {{Thumbs up or thumbs
  down?: semantic orientation applied to unsupervised classification of
  reviews}}.
\newblock In \emph{Proceedings of the 40th Annual Meeting on Association for
  Computational Linguistics}, ACL '02, pages 417--424, Stroudsburg, PA, USA.
  ACL.

\bibitem[{Vilares et~al.()Vilares, Alonso, and
  G{\'{o}}mez-Rodr{\'{i}}guez}]{VilAloGom2015}
D.~Vilares, M.~A. Alonso, and C.~G{\'{o}}mez-Rodr{\'{i}}guez.
\newblock \href
  {https://www.cambridge.org/core/journals/natural-language-engineering/article/a-syntactic-approach-for-opinion-mining-on-spanish-reviews/3CD8A41577BB65402C8B019A56383890}
  {{A syntactic approach for opinion mining on Spanish reviews}}.
\newblock \emph{Natural Language Engineering}, 21(01):139--163.

\bibitem[{Vilares et~al.(2015)Vilares, Alonso, and
  G\'{o}mez-Rodr\'{i}guez}]{VilAloGom2015d}
D.~Vilares, M.~A. Alonso, and C.~G\'{o}mez-Rodr\'{i}guez. 2015.
\newblock \href {https://doi.org/10.1002/asi.23284} {On the usefulness of
  lexical and syntactic processing in polarity classification of {Twitter}
  messages}.
\newblock \emph{Journal of the Association for Information Science and
  Technology}, 66(9):1799--1816.

\bibitem[{Vilares et~al.(2016)Vilares, G{\'{o}}mez-Rodr{\'{i}}guez, and
  Alonso}]{VilGomAloACL2016}
D.~Vilares, C.~G{\'{o}}mez-Rodr{\'{i}}guez, and M.~A. Alonso. 2016.
\newblock \href {http://anthology.aclweb.org/P16-2069} {{One model, two
  languages: training bilingual parsers with harmonized treebanks}}.
\newblock In \emph{Proceedings of the 54th Annual Meeting of the Association
  for Computational Linguistics (Volume 2: Short Papers)}, pages 425--431,
  Berlin, Germany. Association for Computational Linguistics.

\bibitem[{Vilares et~al.(2017)Vilares, Gómez-Rodríguez, and
  Alonso}]{VilGomAloArxiv2016}
David Vilares, Carlos Gómez-Rodríguez, and Miguel~A. Alonso. 2017.
\newblock \href {https://doi.org/10.1016/j.knosys.2016.11.014} {Universal,
  unsupervised (rule-based), uncovered sentiment analysis}.
\newblock \emph{Knowledge-Based Systems}, 118:45--55.

\bibitem[{Villena-Román et~al.(2013)Villena-Román, Lana-Serrano,
  Martínez-Cámara, and González~C.}]{TASS-2012}
J.~Villena-Román, S.~Lana-Serrano, E.~Martínez-Cámara, and J~C González~C.
  2013.
\newblock \href
  {http://journal.sepln.org/sepln/ojs/ojs/index.php/pln/article/view/4657}
  {{TASS - Worshop on Sentiment Analysis at SEPLN}}.
\newblock \emph{Procesamiento de Lenguaje Natural}, 50:37--44.

\bibitem[{Vo and Zhang(2016)}]{vo-zhang:2016:P16-2}
D.~T. Vo and Y.~Zhang. 2016.
\newblock \href {http://anthology.aclweb.org/P16-2036} {Don't count, predict!
  an automatic approach to learning sentiment lexicons for short text}.
\newblock In \emph{Proceedings of the 54th Annual Meeting of the Association
  for Computational Linguistics (Volume 2: Short Papers)}, pages 219--224,
  Berlin, Germany. Association for Computational Linguistics.

\end{thebibliography}



\end{document}